\documentclass[runningheads, envcountsame, a4paper]{llncs}
\usepackage{microtype}
\usepackage[misc]{ifsym}
\usepackage[pdftex]{graphicx}
\usepackage{epstopdf}
\usepackage{enumitem}
\usepackage[]{color}
\makeatletter
\def\maxwidth{
  \ifdim\Gin@nat@width>\linewidth
    \linewidth
  \else
    \Gin@nat@width
  \fi
}
\makeatother

\definecolor{fgcolor}{rgb}{0.345, 0.345, 0.345}

\usepackage{framed}


\definecolor{shadecolor}{rgb}{.97, .97, .97}
\definecolor{messagecolor}{rgb}{0, 0, 0}
\definecolor{warningcolor}{rgb}{1, 0, 1}
\definecolor{errorcolor}{rgb}{1, 0, 0}
\newenvironment{knitrout}{}{}

\usepackage{alltt}

\usepackage{amsmath,amssymb,array}
\usepackage{xcolor}
\usepackage{graphicx}
\usepackage{xargs}
\usepackage{booktabs}
\usepackage{todonotes}
\usepackage[square,numbers,sort&compress,sectionbib]{natbib}
\renewcommand\citet{\cite}

\usepackage[hyphens]{url}
\usepackage{colortbl}
\usepackage{csquotes}
\usepackage{mathtools}
\usepackage{float}
\usepackage{adjustbox}
\usepackage{dsfont}
\usepackage[ruled, vlined, linesnumbered]{algorithm2e}
\let\oldnl\nl
\newcommand{\nonl}{\renewcommand{\nl}{\let\nl\oldnl}}
\usepackage{cleveref}
\usepackage{autonum}

\renewcommand{\S}{S}
\newcommand{\Spi}[1][j]{B_{#1} (\pi)}

\newcommand{\Sc}{C}
\newcommand{\Sj}{\S \cup \{j\}}
\newcommand{\pairs}[1]{\{ (#1) \}}

\newcommand{\Lswitch}[1][\Xperm_{\S}]{L (\fh (#1, \XC ), Y)}
\newcommand{\Eswitch}{\E (\Lswitch)}
\newcommand{\Eorig}{\E (L (\fh (X), Y)) }

\newcommand{\R}{\mathds{R}}

\newcommand{\E}{\mathds{E}}

\newcommand{\iid}{\overset{i.i.d}{\sim}}

\newcommand{\Xspace}[1][P]{\mathcal{X}_{#1}}
\newcommand{\Yspace}{\mathcal{Y}}

\newcommand{\xivec}{(x^{(i)}_1, \ldots, x^{(i)}_p)^\top}
\renewcommand{\xi}[1][(i)]{\mathbf{x}^{#1}}
\newcommand{\xis}[1][(i)]{\mathbf{x}_{\S}^{#1}}

\newcommand{\xic}[1][(i)]{\mathbf{x}_{\Sc}^{#1}}

\newcommand{\yi}[1][i]{y^{(#1)}}

\newcommand{\D}{\mathcal{D}}
\newcommand{\Dset}[1][n]{\pairs{ \mathbf{x}^{(i)}, y^{(i)} }_{i=1}^{#1}}

\newcommand{\Dtest}{\D}
\newcommand{\xj}{\mathbf{x}_j}

\newcommand{\Xperm}{\tilde{X}}
\newcommand{\XC}{X_{\Sc}}
\newcommand{\XS}{X_{\S}}
\newcommand{\xjvec}{(x^{(1)}_j, \ldots, x^{(n)}_j)^\top}

\newcommand{\fh}{\hat{f}}
\newcommand{\fhS}{\fh_{\S}}
\newcommand{\fS}{f_{\S}}
\newcommand{\fhSi}{\fhS^{(i)}}

\newcommand{\MR}{PFI}
\newcommand{\hMR}{\widehat{\MR}}
\newcommand{\MRS}[1][\S]{\MR_{#1}}
\newcommand{\hMRS}[1][\S]{\hMR_{#1}}
\newcommand{\hMRSa}[1][\S]{\hMR_{#1, \text{approx}}}
\newcommand{\w}[1][(\S)]{v_{GE}#1}

\newcommand{\Lii}{L(\fh(\xi), \yi)}
\newcommandx{\Li}[3][1=(k), 2=(i), 3=(i)]{L(\fh(\xis[#1], \xic[#2]), y^{#3})}

\newcommand{\mfeat}{m_{\textnormal{feat}}}
\newcommand{\mobs}{m_{\textnormal{obs}}}

\newcommand{\hPI}{\widehat{PI}_{\S}}

\newcommand{\ICI}{\hMRS^{(i)}}

\newcommandx{\PFIikS}[2][1=(i), 2=(k)]{\Delta L^{#1} (\xis[#2])}

\newcommand{\GEh}[1]{\widehat{GE}_{#1}}

\makeatletter
\newcommand*{\overtabline}{
  \noalign{
    \vskip-.5\dimexpr\ht\@arstrutbox+\dp\@arstrutbox\relax
    \vskip-.2pt\relax
    \hrule
    \vskip-.2pt\relax
    \vskip+.5\dimexpr\ht\@arstrutbox+\dp\@arstrutbox\relax
  }
}

\newcounter{mycomment}

\IfFileExists{upquote.sty}{\usepackage{upquote}}{}
\begin{document}

\title{Visualizing the Feature Importance for Black Box Models}
\titlerunning{Visualizing the Feature Importance for Black Box Models}
\toctitle{Visualizing the Feature Importance for Black Box Models}

\author{Giuseppe~Casalicchio~(\Letter)~\and~Christoph~Molnar~\and~Bernd~Bischl}
\authorrunning{G. Casalicchio et al.}
\tocauthor{Giuseppe~Casalicchio,~Christoph~Molnar,~and~Bernd~Bischl}

\institute{Department of Statistics \\
  Ludwig-Maximilians-University Munich \\
  Ludwigstra{\ss}e 33, 80539 Munich, Germany \\
  \email{giuseppe.casalicchio@stat.uni-muenchen.de}
}

\maketitle

\begin{abstract}
In recent years, a large amount of model-agnostic methods to improve the transparency, trustability, and interpretability of machine learning models have been developed.
Based on a recent method for model-agnostic global feature importance, we introduce a local feature importance measure for individual observations and propose two visual tools: partial importance (PI) and individual conditional importance (ICI) plots which visualize how changes in a feature affect the model performance on average, as well as for individual observations.
Our proposed methods are related to partial dependence (PD) and individual conditional expectation (ICE) plots, but visualize the expected (conditional) feature importance instead of the expected (conditional) prediction.
Furthermore, we show that averaging ICI curves across observations yields a PI curve, and integrating the PI curve with respect to the distribution of the considered feature results in the global feature importance.
Another contribution of our paper is the Shapley feature importance, which fairly distributes the overall performance of a model among the features according to the marginal contributions and which can be used to compare the feature importance across different models.
\end{abstract}

\section{Introduction and Related Work}
\label{sec:introduction}

Machine learning (ML) algorithms such as neural networks and support vector machines (SVM) are often considered to produce black box models because they do not provide any direct explanation for their predictions.
However, these methods often outperform simple linear models or decision trees in predictive performance as they can model complex relationships in the data.
Nevertheless, such simple models are still preferred in areas such as life sciences and social sciences due to their simplicity and interpretability \citep{Lipton2016}.
Many researchers have therefore developed and implemented several model-agnostic interpretability tools, which quantify or visualize feature effects or feature importance \citep{goldstein2015peeking, molnar2018iml, Fisher2018}.

In our context, the terms \emph{feature effect}, feature contribution and feature attribution describe how or to what extent each feature contributes to the \emph{prediction} of the model, either on a local or a global level.
Methods for feature effects include partial dependence (PD) plots \citep{friedman2001greedy}, individual conditional expectation (ICE) plots \citep{goldstein2015peeking} and, more recently, SHAP values \citep{lundberg2018consistent}.
These methods visualize or quantify the relationship and contribution of each feature to the prediction of a model without requiring knowledge about the true values of the target variable.
A method that measures feature effects based on the Shapley value \citep{shapley1953value} from coalitional game theory was first presented for classification in \citep{kononenko2010efficient} and has been extended to regression and global analysis in \citep{vstrumbelj2011general}.
Further developments, visualizations, and generalizations were introduced by \citet{lundberg2017unified, lundberg2018consistent}.
Similar work proposing a general notion of a quantity of interest for the characteristic function of the Shapley value and focusing on the joint and marginal contributions of feature sets was introduced by \citet{Datta2016}.

In biomedical research, for example, measuring the effects of biomedical markers w.r.t. model prediction is as essential as measuring their added value regarding model performance \cite{casalicchio2016residual}.
We use the term \emph{feature importance}\footnote{In the literature, the term feature importance is sometimes also used for methods that only work with model predictions.
In our context, however, we would categorize them under feature effects as they do not take into account the model performance.} to describe how important the feature was for the \emph{predictive performance} of the model, regardless of the shape (e.g., linear or nonlinear relationship) or direction of the feature effect.
This implies that measures of feature importance require knowledge of the true values of the target variable.
The most prominent approach is the permutation importance introduced by Breiman \citep{Breiman2001} for random forests.
It computes the drop in out-of-bag performance after permuting the values of a feature.
A model-agnostic global permutation-based feature importance (PFI) was recently introduced in \citet{Fisher2018}.

\textit{Contributions:} We review model-agnostic global PFI and propose an efficient approximation based on Monte-Carlo integration.
We then introduce a local version of the global PFI, which measures the feature importance of individual observations.
We provide visualizations for local and global PFI, which illustrate how changes in the considered feature affect model performance.
We also relate our new visual tools to PD plots, ICE plots and show that the integral of our PI curve results in the global PFI measure.
Furthermore, we propose a permutation-based Shapley feature importance (SFIMP) measure that fairly distributes the model performance among features and allows the comparison of feature importances across different models.

\section{Preliminaries and Background on Feature Effects}
In this section, we introduce the notation and describe methods focusing on feature effects, which we transfer to feature importance in Section \ref{sec:feature-importance-plots} and \ref{sec:shapley-importance}.

\emph{General Notation:} Consider a $p-$dimensional feature space $\Xspace = (\Xspace[1] \times \hdots \times \Xspace[p])$ with the feature index set $P = \{1, \hdots, p\}$ and a target space $\Yspace$.
Suppose that there is an unknown functional relationship $f$ between $\Xspace$ and $\Yspace$.
ML algorithms try to learn this relationship using training data with observations that have been drawn i.i.d. from an unknown probability distribution $\mathcal{P}$ on the joint space $\Xspace \times \Yspace$.
We consider an arbitrary prediction model $\fh$, fitted on some training data to approximate $f$ and analyze it with model-agnostic interpretability methods.
Let $\D = \Dset$ be a test data set sampled i.i.d. from $\mathcal{P}$ where $n$ is the number of observations in the test set.
We denote the corresponding random variables generated from the feature space by $X = (X_1, \hdots, X_p)$ and the random variable generated from the target space by $Y$.
In our notation, the vector $\xi = \xivec \in \Xspace$ refers to the $i$-th observation, which is associated with the target variable $\yi \in \Yspace$, and $\xj = \xjvec$ denotes the realizations of the $j$-th feature.
We denote the generalization error of a fitted model, which is measured by a loss function $L$ on unseen test data from $\mathcal{P}$, by $GE(\fh, \mathcal{P}) = \E(L(\fh(X), Y))$.
It can be estimated using the test data $\D$ by
\begin{equation}
\label{eq:geest}
\textstyle \GEh{}(\fh, \D) = \frac 1 n \sum_{i=1}^n \Lii.
\end{equation}
A better estimate for the generalization error of an ML algorithm can be obtained using resampling techniques such as cross-validation or bootstrap \citep{Bischl2012}.

\begin{figure}[h]
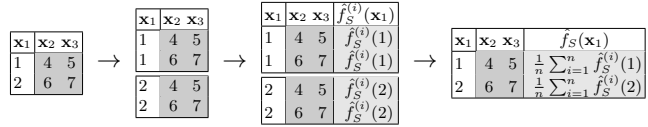

\centering
\scalebox{.7}{
\begin{tabular}{|l|ll|}
\hline
$\mathbf{x}_1$ & $ \mathbf{x}_2$ & $ \mathbf{x}_3$ \\
\hline
1 & \cellcolor[gray]{.8} 4 & \cellcolor[gray]{.8} 5\\
2 & \cellcolor[gray]{.8} 6 & \cellcolor[gray]{.8} 7\\
\hline
\end{tabular}
}
$\rightarrow$
\scalebox{.7}{
\begin{tabular}{|l|ll|}
\hline
$ \mathbf{x}_1$ & $ \mathbf{x}_2$ & $ \mathbf{x}_3$ \\
\hline
1 & \cellcolor[gray]{.8} 4 & \cellcolor[gray]{.8} 5\\
1 & \cellcolor[gray]{.8} 6 & \cellcolor[gray]{.8} 7\\
\hline
\hline
2 & \cellcolor[gray]{.8} 4 & \cellcolor[gray]{.8} 5\\
2 & \cellcolor[gray]{.8} 6 & \cellcolor[gray]{.8} 7\\
\hline
\end{tabular}
}
$\rightarrow$
\scalebox{.7}{
\begin{tabular}{|l|ll|c|}
\hline
$ \mathbf{x}_1$ & $ \mathbf{x}_2$ & $ \mathbf{x}_3$ & $\fhSi (\mathbf{x}_1)$ \\
\hline
1 & \cellcolor[gray]{.8} 4 & \cellcolor[gray]{.8} 5 & \cellcolor{black!10} $\fhSi(1)$  \\
1 & \cellcolor[gray]{.8} 6 & \cellcolor[gray]{.8} 7 & \cellcolor{black!10} $\fhSi(1)$  \\
\hline
\hline
2 & \cellcolor[gray]{.8} 4 & \cellcolor[gray]{.8} 5 & \cellcolor{black!10} $\fhSi(2)$  \\
2 & \cellcolor[gray]{.8} 6 & \cellcolor[gray]{.8} 7 & \cellcolor{black!10} $\fhSi(2)$  \\
\hline
\end{tabular}
}
$\rightarrow$
\scalebox{.7}{
\begin{tabular}{|l|ll|c|}
\hline
$ \mathbf{x}_1$ & $ \mathbf{x}_2$ & $ \mathbf{x}_3$ & $\fhS (\mathbf{x}_1)$ \\
\hline
1 & \cellcolor[gray]{.8} 4 & \cellcolor[gray]{.8} 5 & \cellcolor{black!10}  $\frac{1}{n} \sum_{i=1}^n\fhSi(1)$ \\
2 & \cellcolor[gray]{.8} 6 & \cellcolor[gray]{.8} 7 & \cellcolor{black!10}  $\frac{1}{n} \sum_{i=1}^n\fhSi(2)$ \\
\hline
\end{tabular}
}
\caption{PD plot for an example with $n=2$, $p=3$ and $\S = \{1\}$ and $\Sc = \{2, 3\}$ (marginal effect of $\mathbf{x}_1$ on $\fh$).
We construct a grid using each observed value from $\mathbf{x}_1$, i.e., ${{x}_{1}}^{(1)} = 1$ and ${{x}_{1}}^{(2)} = 2$, and compute the PD function using these grid points.}
\label{eq:pdpscheme}
\end{figure}

\emph{PD Plots \citep{friedman2001greedy}} visualize the marginal relationship between features of interest and the expected prediction of a fitted model on a global level.
Consider a subset of feature indices $\S \subseteq P$ and its complement $\Sc$.
Each observation $\xi[] \in \Xspace$ can be partitioned into $\xis[] \in \Xspace[\S]$ and $\xic[] \in \Xspace[\Sc]$ containing only features from $\S$ and $\Sc$, respectively.
Let $X_S$ and $\XC$ be the corresponding random variables and let the prediction function using features in $\S$, marginalized over features in $\Sc$ be the PD function defined by
$\fS (\xis[]) = \E_{\XC} (\fh(\xis[], \XC) ).$
This definition also covers $f_{\emptyset}(\mathbf{x}_{\emptyset})$ and results in a constant, the average prediction over $\mathcal{P}$.
We can estimate the PD function using Monte-Carlo integration by averaging over feature values $\xic$ in order to marginalize out features in $\Sc$:
\begin{equation}
\label{eq:pdp}
\textstyle \fhS(\xis[]) = \frac{1}{n} \sum_{i = 1}^n \fhSi(\xis[]) = \frac{1}{n} \sum_{i = 1}^n \fh(\xis[], \xic).
\end{equation}
Here, $\fhSi(\xis[]) = \fh(\xis[], \xic)$ can be read in two ways: a) the prediction of the $i$-th observation with replaced feature values in $\S$ taken from $\xi[]$ or b) the prediction of $\xi[]$ with replaced values in $\Sc$ taken from the $i$-th observation.
Plotting the pairs $\pairs{\xis[*^{(k)}], \fhS(\xis[*^{(k)}])}_{k=1}^m$ using (often $m < n$) grid points denoted by $\xis[*^{(1)}], \hdots, \xis[*^{(m)}]$ yields a PD curve.
Fig. \ref{eq:pdpscheme} illustrates the PD principle for a simple example.

\emph{ICE Plots \citep{goldstein2015peeking}:} The averaging in Eq. \eqref{eq:pdp} of the PD function can obfuscate more complex relationships resulting from feature interactions, i.e. when the partial relationship of one or more observations depends on other features.
ICE plots address this problem by visualizing to what extent the prediction of a single observation changes when the value of the considered feature changes.
Instead of plotting the pairs $\pairs{\xis[*^{(k)}], \fhS(\xis[*^{(k)}])}_{k=1}^m$, ICE plots visualize the pairs $\pairs{\xis[*^{(k)}], \fhSi(\xis[*^{(k)}])}_{k=1}^m$ for each observation indexed by $i \in \{1, \hdots, n\}$.

\emph{Shapley Value:} A coalitional game is defined by a set of players $P$, which can form coalitions $S \subseteq P$. Each coalition $S$ achieves a certain payout.
The characteristic function $v: 2^P \rightarrow \R$ maps all $2^p$ possible coalitions to their payouts.
The Shapley value \citep{shapley1953value} now fairly assigns a value to each player depending on their contribution in all possible coalitions.
This concept was transferred to feature effect estimation in \citet{kononenko2010efficient}.
We could explain the prediction of a single, fixed observation $\xi[]$ by regarding features as players, who form various coalitions (subsets) $\S$ to achieve the prediction $\fh(\xi[])$. For each coalition $S$, we are only allowed to access values of features from $\S$.
A natural definition of the payout is the PD value $\fS (\xis[])$, which we shift so that an
empty set of no features is assigned a value of $0$ -- which is required by the general Shapley value definition:
\begin{equation}
\label{eq:valuefunction}
v(\xis[]) = \E_{\XC} (\fh(\xis[], \XC) ) - \E_X(\fh(X)) = \fS (\xis[]) - f_{\emptyset}(\mathbf{x}_{\emptyset}).
\end{equation}
The marginal contribution of feature $j$, joining a coalition $\S$, is defined as
\begin{equation}
\label{eq:mc}
\Delta_j (\xis[]) = v (\xi[]_{\Sj}) - v (\xis[]) = f_{\Sj} (\xi[]_{\Sj}) - f_{\S} (\xis[]).
\end{equation}

Let $\Pi$ be the set of all possible permutations over the index set $P$.
For a permutation $\pi \in \Pi$, we denote the set of features that are in order \textit{before} feature $j$ as $\Spi$.
For example, for $p=4$, if we consider feature $j=4$ and permutation $\pi = \{2, 3, \textbf{4}, 1\}$, then $\Spi[4] = \{2, 3\}$.
For an observation $\xi[]$ and its feature value for feature $j$, the Shapley value can be estimated by
\begin{equation}
\begin{array}{r@{}l}
\hat{\phi}_j (\xi[]) &{}=
\frac {1} {p!} \sum_{\pi \in \Pi} \hat{\Delta}_j ( \xi[]_{\Spi}) \\
&{}= \frac {1} {p!} \sum_{\pi \in \Pi} \fh_{\Spi \cup \{j\}} (\xi[]_{\Spi \cup \{j\} }) - \fh_{\Spi} (\xi[]_{\Spi}) \\
&{}= \frac {1} {p! \cdot n} \sum_{\pi \in \Pi}
\sum_{i = 1}^n \fh_{\Spi \cup \{j\}}^{(i)} (\xi[]_{\Spi \cup \{j\} }) - \fh_{\Spi}^{(i)} (\xi[]_{\Spi}),
\end{array}
\end{equation}
where $\fh_{\Spi}$ and $\fh_{\Spi \cup \{j\}}$ are estimated by Eq. \eqref{eq:pdp}.
An efficient approximation based on Monte-Carlo integration using $m$ rather than $p! \cdot n$ summands was proposed by \citet{vstrumbelj2011general}.
Consider the following example to illustrate the Shapley value:
The features enter a room in a random order specified by the permutation $\pi$.
All features in the room participate in the game, i.e., they contribute to the model prediction.
The Shapley value $\phi_j$ is the average additional contribution of feature $j$ by joining whatever features already entered the room before.

\section{Permutation-based Feature Importance}
\label{sec:feature-importance}

\paragraph{Background.}
The permutation importance for random forests introduced in \citet{Breiman2001} measures the performance, e.g., the mean squared error (MSE), of each tree within a random forest using out-of-bag samples.
The performance is measured once with and once without permuted values of the feature of interest.
The difference between those two performance values is computed for each tree and averaged to yield the feature importance.
Permuting the values of a feature breaks the association between the feature and the target variable and results in a large drop in performance if the considered feature is important.
A model-agnostic global PFI for features included in $\S$ can be defined as
\begin{equation}
\label{eq:pfi2}
\begin{array}{r@{}l}
\MRS
&{}= \Eswitch - \Eorig
\end{array}
\end{equation}
where $\Xperm_\S$ refers to an independent replication of $\XS$, which is also independent of $\XC$ and $Y$.
This implies that $\Xperm_\S$ is a new (multivariate) random variable, which is distributed as
$\XS$, but independent of everything else.
This definition is analogous to the permutation-based model reliance introduced by \citep{Fisher2018} and relates to the definition in \citet{Gregorutti2017} where the authors focus on random forests.
The larger the value of $\MRS$, the more substantial the increase in error when we permute feature values in $S$, and the more important we deem the feature set $S$.
According to \citep{Fisher2018}, the use of the ratio
$\MRS = \Eswitch / \Eorig$
instead of the difference might be more comparable across different models, as it always refers to the relative drop in performance with respect to the standard generalization error.
However,
using the ratio can result in numerically unstable estimations if the denominator is close or equal to zero.
Thus, both definitions have drawbacks that we try to are address in Section \ref{sec:shapley-importance}.

\paragraph{Estimating and Approximating the PFI.}
The first term of Eq. \eqref{eq:pfi2} encodes the expected generalization error under perturbation of features in feature set $S$, which can be formulated as:
\begin{equation}
\label{eq:ge}
\begin{array}{lcl}
\Eswitch
&=& \E_{(\XC,Y)} (\E_{\Xperm_{\S}|(\XC,Y)} ( \Lswitch )) \\
&=& \E_{(\XC,Y)} (\E_{\Xperm_{\S}} ( \Lswitch )) \\
&=& \E_{(\XC,Y)} (\E_{\XS} ( \Lswitch[\XS] ))
\end{array}
\end{equation}
In the derivation above, the first equality follows from the \enquote{law of total expectation}, the second from the independence of $\Xperm_{\S}$ from $(\XC,Y)$, and the third because $\Xperm_{\S}$ is distributed as $\XS$.
We can plug in an estimator for the inner expected value and denote the estimate of this quantity by
\begin{equation}
\label{eq:ge2}
\begin{array}{r@{}l}
\widehat{GE}_{\Sc}(\fh, \Dtest)
&{}= \frac{1}{n} \sum_{\substack{i = 1}}^n \frac{1}{n} \sum_{\substack{k = 1}}^n \Li.
\end{array}
\end{equation}
The index $\Sc$ in $GE_{\Sc}$ emphasizes that the set of features in $\Sc$ were not replaced with a perturbed random variable and can thus be seen as the model performance using features in $\Sc$ (and ignoring those in $\S$).
The above estimator is analogous to the V-statistic \citep{serfling2009approximation} and may also be replaced by the unbiased U-statistic using $\frac{1}{n} \sum_{\substack{i = 1}}^n \frac{1}{n-1} \sum_{\substack{k \neq i}} \Li$ as proposed by \citep{Fisher2018}.\footnote{For the sake of simplicity, we consider the V-statistic throughout the article.
However, all calculations and approximations based on Eq. \eqref{eq:ge2} still apply -- with slight modifications -- when using the U-statistic.}
The estimator scales with $O(n^2)$ (for a given set $C$, and assuming $\fh$ can be computed in constant time), which can be expensive when $n$ is large.
However, we can use a different formulation to motivate an approximation for Eq. \eqref{eq:ge2}:
Let $\{\pmb{\tau}_1, \hdots, \pmb{\tau}_{n!}\}$ be the set of all possible permutation vectors over the observation index set $\{1, \hdots, n\}$.
As shown by \citep{Fisher2018}, we can replace Eq. \eqref{eq:ge2} by the equivalent formulation
\begin{equation}
\label{eq:ge3}
\begin{array}{r@{}l}
\widehat{GE}_{\Sc, \text{perm}}(\fh, \Dtest)
&{}= \frac{1}{n} \sum_{\substack{i = 1}}^n \frac{1}{n!} \sum_{\substack{k = 1}}^{n!} \Li[(\tau_k^{(i)})].
\end{array}
\end{equation}
If we approximate $\widehat{GE}_{\Sc, \text{perm}}$ by Monte-Carlo integration using only $m$ randomly selected permutations rather than all $n!$ permutations, we obtain
\begin{equation}
\label{eq:ge4}
\begin{array}{r@{}l}
\widehat{GE}_{\Sc, \text{approx}}(\fh, \Dtest)
&{}= \frac{1}{n} \sum_{\substack{i = 1}}^n \frac{1}{m} \sum_{\substack{k = 1}}^{m} \Li[(\tau_k^{(i)})].
\end{array}
\end{equation}
The approximation refers to permuting features in $\S$ repeatedly (i.e., $m$ times) and averaging the resulting model performances.\footnote{By the same logic, we could also directly approximate Eq. \eqref{eq:ge2} by summing over $m$ randomly selected feature values for features in $\S$ instead of using all of them. We here opted for Eq. \eqref{eq:ge4}, due to the in our opinion interesting relation to the random forest permutation importance explained at the end of this section.}
The PFI from Eq. \eqref{eq:pfi2} can be estimated using Eq. \eqref{eq:ge2} for the first term and using Eq. \eqref{eq:geest} for the last term.
Including the summands into an iterated sum yields the estimate
\begin{equation}
\label{eq:pfiest}
\begin{array}{r@{}l}
\hMRS
&{}= \frac{1}{n^2} \sum_{i = 1}^n \sum_{k = 1}^n \left( \Li - \Lii \right).
\end{array}
\end{equation}
If we use Eq. \eqref{eq:ge4} rather than Eq. \eqref{eq:ge2}, we obtain the approximation
\begin{equation}
\label{eq:pfiapprox}
\resizebox{.925 \textwidth}{!}{
$\begin{array}{r@{}l}
\hMRSa
&{}= \frac{1}{n \cdot m} \sum_{i = 1}^n \sum_{k = 1}^m \left(\Li[(\tau_k^{(i)})] - \Lii \right).
\end{array}$
}
\end{equation}
Eq. \eqref{eq:pfiapprox} is identical to the permutation importance of random forests formalized in \citet{Gregorutti2017} if we consider $m$ as the number of trees, replace $n$ with the number of out-of-bag samples per tree and replace the model $\fh$ with the individual trees fitted within a random forest, i.e., $\fh_k$.

\section{Visualizing Global and Local Feature Importance}
\label{sec:feature-importance-plots}

Consider the summands in Eq. \eqref{eq:pfiest} and denote them by
\begin{equation}
\PFIikS[(i)][] = \Li[] - \Lii.
\end{equation}
This quantity refers to the change in performance between the $i$-th observation with and without replaced feature values $\xis[]$.
Inspired by ICE plots, we introduce \textit{individual conditional importance} (ICI) plots which visualize the pairs $\pairs{ \xis[{(k)}], \PFIikS }_{k = 1}^{n}$ for all observations $i = 1, \hdots, n$.
We define the local feature importance of the $i$-th observation (regarding features in $\S$) as the integral of the corresponding ICI curve with respect to the distribution of $\XS$.
It is estimated by
$\textstyle \ICI = \frac {1}{n} \sum_{k = 1}^n \PFIikS$
and can be interpreted as the expected change in performance of the $i$-th observation after marginalizing its features in $\S$.
It also refers to the contribution of the $i$-th observation to the global PFI (see later in Eq. \eqref{ref:relationpfi}).
To the best of our knowledge, a similar definition for local feature importance only exists in the context of random forests, e.g., in \citep{Cutler2012}.

Analogous to the PD function from Eq. \eqref{eq:pdp}, we introduce the \textit{partial importance} (PI) function as the expected change in performance at a specific value $\xis[]$, which can be estimated by
$\hPI (\xis[]) = \frac {1}{n} \sum_{i = 1}^n \PFIikS[(i)][]$.
Consequently, a PI plot visualizes the pairs $\pairs{ \xis[{(k)}], \hPI (\xis[(k)])}_{k = 1}^n$ and refers to the pointwise average of all ICI curves across all observations at fixed grid points $\xis[]$.

\begin{figure}[h]
\scalebox{.7}{
\begin{tabular}{|l|ll|}
\hline
$\mathbf{x}_1$ & $ \mathbf{x}_2$ & $ \mathbf{x}_3$ \\
\hline
1 & \cellcolor[gray]{.8} 4 & \cellcolor[gray]{.8} 5\\
2 & \cellcolor[gray]{.8} 6 & \cellcolor[gray]{.8} 7\\
3 & \cellcolor[gray]{.8} 8 & \cellcolor[gray]{.8} 9\\
\hline
\end{tabular}
\hspace{-8pt}
}
$\xrightarrow[]{\text{a)}}$
\hspace{-3pt}
\scalebox{.7}{
\begin{tabular}{|c|cc|}
\hline
$ \mathbf{x}_1$ & $ \mathbf{x}_2$ & $ \mathbf{x}_3$ \\
\hline
1 & \cellcolor[gray]{.8} 4 & \cellcolor[gray]{.8} 5\\
1 & \cellcolor[gray]{.8} 6 & \cellcolor[gray]{.8} 7\\
1 & \cellcolor[gray]{.8} 8 & \cellcolor[gray]{.8} 9\\
\hline
\hline
2 & \cellcolor[gray]{.8} 4 & \cellcolor[gray]{.8} 5\\
2 & \cellcolor[gray]{.8} 6 & \cellcolor[gray]{.8} 7\\
2 & \cellcolor[gray]{.8} 8 & \cellcolor[gray]{.8} 9\\
\hline
\hline
3 & \cellcolor[gray]{.8} 4 & \cellcolor[gray]{.8} 5\\
3 & \cellcolor[gray]{.8} 6 & \cellcolor[gray]{.8} 7\\
3 & \cellcolor[gray]{.8} 8 & \cellcolor[gray]{.8} 9\\
\hline
\end{tabular}
\hspace{-8pt}
}
$\xrightarrow[]{\text{b)}}$
\hspace{-3pt}
\scalebox{.7}{
\begin{tabular}{l|c|cc|c|}
\hline
i & $ \mathbf{x}_1$ & $ \mathbf{x}_2$ & $ \mathbf{x}_3$ & $\Delta L^{(i)} (\mathbf{x}_1)$\\
\hline
1 & 1 & \cellcolor[gray]{.8} 4 & \cellcolor[gray]{.8} 5 & \cellcolor{black!10} 0  \\
2 & 1 & \cellcolor[gray]{.8} 6 & \cellcolor[gray]{.8} 7 & \cellcolor{black!10} 0.6 \\
3 & 1 & \cellcolor[gray]{.8} 8 & \cellcolor[gray]{.8} 9 & \cellcolor{black!10} 0.3 \\
\hline
\hline
1 & 2 & \cellcolor[gray]{.8} 4 & \cellcolor[gray]{.8} 5 & \cellcolor{black!10} 0.65  \\
2 & 2 & \cellcolor[gray]{.8} 6 & \cellcolor[gray]{.8} 7 & \cellcolor{black!10} 0 \\
3 & 2 & \cellcolor[gray]{.8} 8 & \cellcolor[gray]{.8} 9 & \cellcolor{black!10} 0.25 \\
\hline
\hline
1 & 3 & \cellcolor[gray]{.8} 4 & \cellcolor[gray]{.8} 5 & \cellcolor{black!10} 0.7  \\
2 & 3 & \cellcolor[gray]{.8} 6 & \cellcolor[gray]{.8} 7 & \cellcolor{black!10} 0.5  \\
3 & 3 & \cellcolor[gray]{.8} 8 & \cellcolor[gray]{.8} 9 & \cellcolor{black!10} 0 \\
\hline
\end{tabular}
\hspace{-8pt}
}
$\xrightarrow[]{\text{c)}}$
\hspace{-3pt}
\scalebox{.7}{
\begin{tabular}{|l|ll|c|}
\hline
$ \mathbf{x}_1$ & $ \mathbf{x}_2$ & $ \mathbf{x}_3$ & $\hPI (\mathbf{x}_1)$ \\
\hline
1 & \cellcolor[gray]{.8} 4 & \cellcolor[gray]{.8} 5 & \cellcolor{black!10}
0.3
\\
2 & \cellcolor[gray]{.8} 6 & \cellcolor[gray]{.8} 7 & \cellcolor{black!10}
0.3
\\
3 & \cellcolor[gray]{.8} 8 & \cellcolor[gray]{.8} 9 & \cellcolor{black!10}
0.4
\\
\hline
\end{tabular}
}
\begin{minipage}{0.325\textwidth}
\begin{knitrout}\small
\definecolor{shadecolor}{rgb}{0.969, 0.969, 0.969}\color{fgcolor}

{\centering \includegraphics[width=1\textwidth]{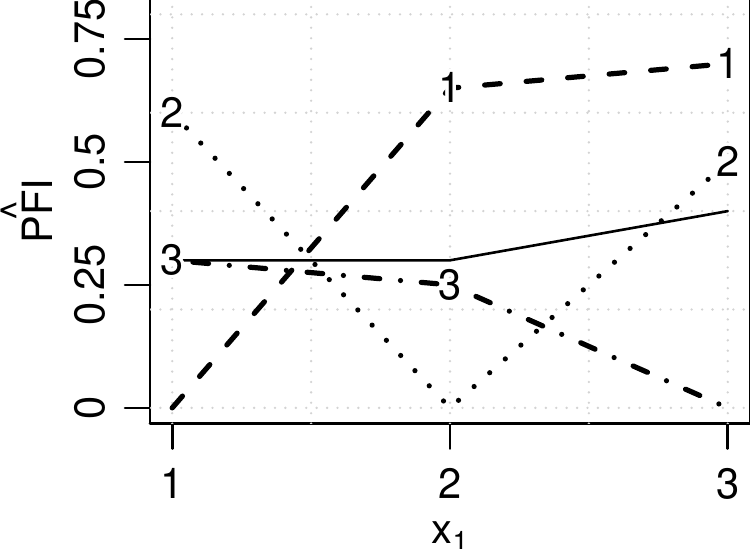}

}

\end{knitrout}
\end{minipage}
\caption{The tables on the left side illustrate the required steps to create ICI curves and PI plots as described in Algorithm \ref{algo:piplot}. The right plot visualizes the ICI curves of individual observations for $i = 1, 2, 3$ (dotted and dashed lines) and the PI curve (solid line) which is the average of ICI curves at each point of the abscissa. All points belonging to the same observation are connected by a line to produce the ICE curves.}
\label{eq:pischeme}
\end{figure}

Fig. \ref{eq:pischeme} illustrates the computation of ICI and PI curves for the first feature.
It also shows the $n$ grid points for which $\PFIikS[(i)][(i)] = 0 \; \forall i$.
We can omit these points by plotting the pairs $\pairs{ \xis[{(k)}], \PFIikS }_{k \in \{1, \hdots, n\} \setminus \{i\}}$ to visualize the unbiased estimation of the feature importance proposed by \citet{Fisher2018}.
Visualizing the ICI curves for the approximation in Eq. \eqref{eq:pfiapprox} implies that some grid points are randomly skipped because the feature values used as grid points are implicitly determined by the randomly selected permutations in Eq. \eqref{eq:pfiapprox}.
The ICI curves, the PI curve, and the global PFI are related:
Averaging all ICI curves pointwise yields a PI curve.
Integrating the PI curve (as well as averaging the integral of all ICI curves) using Monte-Carlo integration over all points $\{\xis[{(k)}]\}_{k=1}^n$ yields an equivalent estimate of the global PFI from Eq. \eqref{eq:pfiest}:
\begin{equation}
\label{ref:relationpfi}
\textstyle \hMRS = \frac 1 n \sum_{i=1}^n \ICI = \frac 1 n \sum_{k=1}^n \hPI (\xis[(k)]).
\end{equation}
We propose to additionally inspect the PI and ICI curves instead of focusing on a single PFI value.
PI curves enable the user to identify regions in which the feature importance is higher or lower than its global PFI.
ICI curves additionally enable the user to identify (suspicious) observations that impact the global PFI strongly and can reveal heterogeneity in the feature importance among the observations, which remain hidden in the PI plots (see also Section \ref{sec:application}).

Algorithm \ref{algo:piplot} describes a procedure for obtaining PI and PD plots, which also allows to return ICI and ICE plots by visualizing $\pairs{ \xis[*^{(k)}], \PFIikS[(i)][*^{(k)}] }_{k = 1}^{m}$ and $\pairs{\xis[*^{(k)}], \fhSi(\xis[*^{(k)}])}_{k=1}^m$ for all observations $i$.
Similar to PD and ICE plots, we can use all $k = 1, \hdots, n$ or a random sample (of size $m < n$) of feature values from $\S$ as grid points for PI and ICI plots.

\begin{algorithm}
\caption{PD plot and PI plot}\label{algo:piplot}
\begin{enumerate}[nosep]
\item Choose $m$ grid points $\xis[*^{(1)}], \hdots, \xis[*^{(m)}]$.
\item Repeat the following steps for the $k$-th grid point:
\begin{itemize}
\item[a)] Modify the data by replacing all observed values in $\mathbf{x}_S$ with the constant values from the $k$-th grid point $\xis[*^{(k)}]$.
\item[b)] Use the modified data from a), the prediction function $\hat{f}$ and the loss function $L$ and calculate for all individual observations:
\begin{itemize}
\item[i)] $\fhSi(\xis[*^{(k)}]) = \hat{f} (\xis[*^{(k)}], \xic)$
\item[ii)]
$\PFIikS[(i)][*^{(k)}] = \Li[*^{(k)}] - \Lii$
\end{itemize}
\item[c)] Aggregate the individual values:
\begin{itemize}
\item[i)]
$\hat{f}_{S}(\xis[*^{(k)}]) = \frac{1}{n} \sum_{i = 1}^n \fhSi(\xis[*^{(k)}])$
\item[ii)]
$\hPI (\xis[*^{(k)}]) = \frac{1}{n} \sum_{i = 1}^n \PFIikS$
\end{itemize}
\end{itemize}
\item Plot the pairs $\pairs{\xis[*^{(k)}], \hat{f}_{S}(\xis[*^{(k)}]) }_{k=1}^m$ and $\pairs{\xis[*^{(k)}], \hPI (\xis[*^{(k)}]}_{k=1}^m.$
\end{enumerate}
\end{algorithm}

\section{Shapley Feature Importance}
\label{sec:shapley-importance}

In this section, we introduce the \textit{S}hapley \textit{F}eature \textit{IMP}ortance (SFIMP) measure, which allows to easily visualize and interpret the contribution of each feature to the model performance.
Our goal is to fairly distribute the performance difference among the individual features between the scenario when all features are used and when all features are ignored, which is illustrated in Fig. \ref{fig:performance}.

\begin{knitrout}\small
\definecolor{shadecolor}{rgb}{0.969, 0.969, 0.969}\color{fgcolor}\begin{figure}

{\centering \includegraphics[width=\maxwidth]{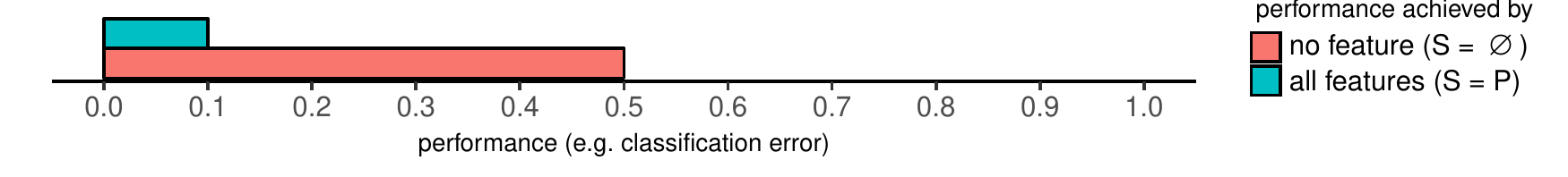}

}

\caption[Illustration of the difference in model performance that we want to fairly distribute among the features]{Illustration of the difference in model performance that we want to fairly distribute among the features. The model performance (e.g., classification error) is $0.1$ when using all features (green bar) and $0.5$ when ignoring all features (red bar). Our goal is to fairly distribute the resulting performance difference of $0.4$ among all involved features based on their marginal contribution.}\label{fig:performance}
\end{figure}

\end{knitrout}

The Shapley value was used in \citet{Cohen2007} for a fair attribution of the difference in model performance.
However, the authors focused on feature selection which requires refitting the model by leaving out or including features.
This can lead to different results of the learning algorithm since different relationships can be learned due to the absence of features. This is reasonable in the context of feature selection.
However, as we measure the feature importance of an already fitted model, we prefer marginalizing over features rather than omitting them completely.
Inspired by Eq. \eqref{eq:valuefunction}, we define the characteristic function of the coalition of features in $\S \subseteq P$ based on Eq. \eqref{eq:ge2} as:
\begin{equation}
\label{eq:performancegain}
\begin{array}{r@{}l}
\w
&{} = \widehat{GE}_{\S} (\fh, \Dtest) - \widehat{GE}_{\emptyset} (\fh, \Dtest).
\end{array}
\end{equation}
The characteristic function measures the change in performance between using features in $\S$ (i.e., ignoring features in its complement $\Sc$ by marginalizing over them) and ignoring all features.
This is similar to Eq. \eqref{eq:pfiest} which, in contrast, measures the change in performance between ignoring features in $\S$ and using all features.
Since the error $\widehat{GE}_{\emptyset} (\fh, \Dtest)$ (no features are considered, i.e., all features are marginalized out) is usually greater than $\widehat{GE}_{\S} (\fh, \Dtest)$, $\w$ will have negative values.\footnote{We prefer the definition in Eq. \eqref{eq:performancegain} as it directly shows the relation to Eq. \eqref{eq:valuefunction}, however, we could also swap the sign as discussed at the end of this section.}
The marginal contribution of a feature $j$ to a coalition of features in $\S$ is given by
\begin{equation}
\label{eq:mc}
\Delta_j (\S)
= \w[(\Sj)] - \w
= \widehat{GE}_{\Sj} (\fh, \Dtest) - \widehat{GE}_{\S} (\fh, \Dtest).
\end{equation}
If we consider a permuted order $\pi \in \Pi$ of the features, where $\Spi$ is the set of features occurring before feature $j$, we obtain the Shapley value estimation
\begin{equation}
\label{eq:shap2}
\begin{array}{r@{}l}
\hat{\phi}_j (\w[])
&{}= \frac {1} {p!} \sum_{\pi \in \Pi} \Delta_j (\Spi) \\
&{}= \frac {1} {p!} \sum_{\pi \in \Pi} \widehat{GE}_{\Spi \cup \{j\}} (\fh, \Dtest) - \widehat{GE}_{\Spi} (\fh, \Dtest),
\end{array}
\end{equation}
which refers to the SFIMP measure of feature $j$.
Computing Eq. \eqref{eq:shap2} is computationally expensive when the number of features $p$ is large, even if we use the approximation of the model performance from Eq. \eqref{eq:ge4}.
We therefore suggest an efficient procedure in Algorithm \ref{algo:shapleyimp}.
The Shapley value satisfies the following four desirable properties as already worked out in \citet{Cohen2007}:
\begin{enumerate}
  \item Efficiency: $\sum_{j=1}^p \phi_j = \w[(P)]$.
  All SFIMP values add up to $\w[(P)]$, i.e., the difference in performance between the scenario when all features are used and when all features are ignored. This allows us to calculate the proportion of explained importance for each feature $j$ using $\frac{\phi_j}{\sum_{j=1}^p \phi_j}$.
  \item Symmetry: If $\w[(\S \cup \{j\})] = \w[(\S \cup \{k\})] $ for all $\S \subseteq \{1, \ldots, p\} \setminus \{j, k\}$, then $\phi_j = \phi_k$.
  Two features $j$ and $k$ have the same SFIMP values if their marginal contribution to all possible coalitions is the same.
  \item Dummy property: If $\w[(\S \cup \{j\})]= \w$ for all $\S \subseteq P$, then $\phi_j = 0$.
  The SFIMP value of a feature $j$ is zero if its marginal contribution does not change no matter to which coalition $\S$ the feature is added.
  \item Additivity: $\phi_j (\w[] + w_{GE}) = \phi_j (\w[]) + \phi_j (w_{GE})$.
  The SFIMP value resulting from a single game with two combined performance measures $\phi_j (\w[] + w_{GE})$ is the same as the sum of the two SFIMP values resulting from two separate games with corresponding characteristic functions, i.e., $\phi_j (\w[]) + \phi_j (w_{GE})$.

  Linearity: $ \phi_j (c \cdot \w[]) = c \cdot \phi_j (\w[])$.
  Any multiplication of the performance measure with a constant $c$ does not affect the feature ranking.
\end{enumerate}

\begin{algorithm}
\DontPrintSemicolon
\KwIn{$\mfeat$, $\mobs$, $\fh$, $L$, $\Dtest = \Dset$}
  \ForAll{$k \in \{1, \hdots, \mfeat\}$}
  {
	choose a random permutation of the feature indices $\pi \in \Pi$.

	set $\S = \Spi$ containing features that won't be permuted.

  set $\widehat{GE}_{\S, \text{perm}} = 0$ and $\widehat{GE}_{\Sj, \text{perm}} = 0$.

	 \ForAll{$l \in \{1, \hdots, \mobs\}$} {
	 	choose a random permutation of observation indices $\pmb{\tau} \in \{\pmb{\tau}_1, \hdots, \pmb{\tau}_{n!}\}$.

	 	measure performance by permuting features w.r.t. $\pmb{\tau} = (\tau^{(1)}, \hdots, \tau^{(n)})$:\newline
	 	$\widehat{GE}_{\S, \text{perm}} = \widehat{GE}_{\S, \text{perm}} + \frac 1 n \sum_{i = 1}^n L(\fh(\xis, \xic[(\tau^{(i)})]), \yi))$
	 	$\widehat{GE}_{\Sj, \text{perm}} = \widehat{GE}_{\Sj, \text{perm}} + \frac 1 n \sum_{i = 1}^n L(\fh(\xi_{\Sj}, \xi[(\tau^{(i)})]_{\Sc \setminus \{j\}}), \yi))$
	 }
	 compute marginal contribution for feature $j$ in iteration $k$: \newline
	 $\Delta_j^{(k)}(\S) = \frac 1 \mobs \cdot (\widehat{GE}_{\Sj, \text{perm}} - \widehat{GE}_{\S, \text{perm}})$
  }
  \Return{$\hat{\phi}_j = \frac 1 \mfeat  \sum_{k = 1}^{\mfeat} \Delta_j^{(k)}(\S)$}
\caption{Approximation of SFIMP values: Contribution of $j$-th feature towards the model performance.}\label{algo:shapleyimp}
\end{algorithm}

The properties above imply that fairly distributing the drop in performance using
$v_{\MR}(\S) = \hMRS = \widehat{GE}_{\Sc}(\fh, \Dtest) - \widehat{GE}_{P}(\fh, \Dtest)$
results in the same Shapley values (except for the sign) and is equivalent to using $-\w[(P)]$.
The SFIMP measure can thus be seen as an extension of the PFI measure in the sense that it additionally fairly distributes the importance values among features.
The PFI measure ignores features in $\S$ by permuting or marginalizing over them, which destroys any correlation and interaction of features in $\Sc$ with features in $\S$.
Consequently, the PFI of a feature also includes the importance of any interaction with that feature and features in $\Sc$ and therefore an interaction will be fully attributed to all involved features.
The SFIMP measure solves this issue as it considers the marginal contribution of a feature and equally distributes the importance of interactions among the interacting features.
This allows comparing feature importances across different models.

\section{Simulations and Application}
\label{sec:application}

For full reproducibility, all our proposed methods are available in the \texttt{R} package \texttt{featureImportance}\footnote{https://github.com/giuseppec/featureImportance.}.
The repository also contains the \texttt{R} code, which is partly based on \texttt{batchtools} \citep{lang2017batchtools}, for the application and simulation in this section.

\subsection{Simulations}
\paragraph{PI and ICI Plots.}

Consider the following data-generating model:
\begin{equation}
Y = X_1 + X_2 + 10 X_1 \cdot \mathds{1}_{X_3 = 0} + 10 X_2 \cdot \mathds{1}_{X_3 = 1} + \epsilon,
\end{equation}
$$X_1, X_2 \iid \mathcal{N}(0, 1), X_3 \sim \mathcal{B}(1, 0.5), \epsilon \sim \mathcal{N}(0, 0.5).$$
We simulate a training data set with $10000$ observations, train a random forest and compute the global PFI on 100 test sets of size $n = 100$ sampled from the same distribution.
We demonstrate that, by merely inspecting the global PFI, the features $X_1$ and $X_2$ would be considered equally important.
However, due to the interactions, it is clear that feature $X_1$ should be considered more important than $X_2$ when $X_3 = 0$ and vice-versa when $X_3 = 1$.

According to Eq. \eqref{ref:relationpfi}, averaging the local feature importances (i.e., the integral of all ICI curves) results in the global PFI.
Having at hand the local feature importance of each observation allows calculating the PFI conditional on other features.
This does not require additional time-consuming calculations, as we only have to average the already computed local feature importances according to the condition considered in the conditional PFI.
The relevance of conditional feature importance in the case of random forests with correlated features was discussed in \citet{Strobl2008}.
In Fig. \ref{fig:iciplot2}, we illustrate the usefulness of a model-agnostic conditional PFI in case of interactions by showing the PI curves of $X_1$ and $X_2$ conditional on the binary feature $X_3$.
The integral of these conditional PI curves refers to the PFI conditional on $X_3$.
Its value differs depending on the two groups introduced by feature $X_3$, which suggests that there is an interaction between the features $X_1$ and $X_3$ as well as $X_2$ and $X_3$.

Table \ref{tab:cimp} shows that feature $X_1$ and $X_2$ are almost equally important if we consider the unconditional global PFI.
However, a different ranking of features is obtained when we compute the PFI conditional on $X_3$.
Thus, inspecting PI and ICI curves conditional on other feature values may help in detecting interactions.

\begin{table}[ht]
\centering
\caption{The mean and the standard deviation (numbers in brackets) of the PFI values estimated using the 100 simulated test data sets.}
\label{tab:cimp}
\begin{tabular}{c|c|c}
  \hline
 & $X_1$ & $X_2$ \\
  \hline
global PFI & 77.976 (14.15) & 76.764 (13.89) \\
  PFI for $X_3 = 0$ & 152.49 (26.06) & 1.428 (1.32) \\
  PFI for $X_3 = 1$ & 1.261 (1.03) & 151.489 (24.69) \\
   \hline
\end{tabular}
\end{table}

\begin{knitrout}\small
\definecolor{shadecolor}{rgb}{0.969, 0.969, 0.969}\color{fgcolor}\begin{figure}[ht]
\includegraphics[width=\maxwidth]{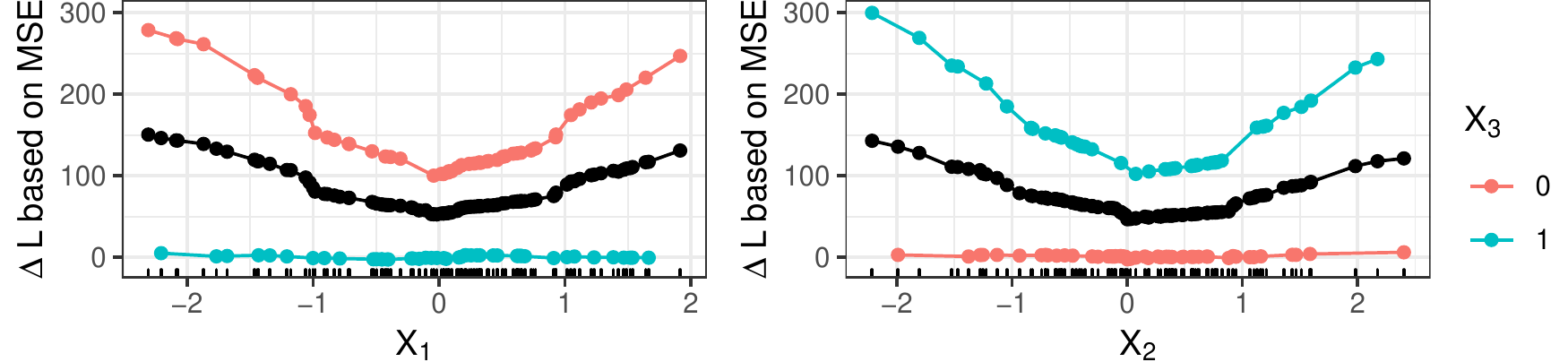} \caption{PI curves of $X_1$ and $X_2$ calculated using all observations (black line) and conditional on $X_3 = 0$ (red line) and $X_3 = 1$ (green line). The points plotted on the lines refer to the observed feature values that were used as grid points to produce the corresponding PI curves as described in Algorithm \ref{algo:piplot}.}\label{fig:iciplot2}
\end{figure}

\end{knitrout}

\paragraph{Shapley Feature Importance.}

We illustrate how the SFIMP measure can be used to compare the feature importance across different models and present the results of a small simulation study to compare the SFIMP measure introduced in Section \ref{sec:shapley-importance} with the difference-based and the ratio-based PFI discussed in Section \ref{sec:feature-importance}.
Consider the following data-generating linear model with a simple interaction:
\begin{equation}
\label{eq:datagenerator}
Y = X_1 + X_2 + X_3 + X_1 \cdot X_2 + \epsilon, \;\;\;\;\; X_1, X_2, X_3 \iid \mathcal{N}(0, 1), \epsilon \sim \mathcal{N}(0, 0.5).
\end{equation}
All three features and the interaction of $X_1$ and $X_2$ have the same linear effect on the target $Y$.
We simulate training data with $10000$ observations and train four learning algorithms using the \texttt{mlr} R package \citet{JMLR:v17:15-066} in their defaults:
An SVM with Gaussian kernel (\texttt{ksvm}), a random forest (\texttt{randomForest}), a simple linear model (\texttt{lm}) and another one that considers 2-way interaction effects (\texttt{rsm}).
We use a test set with $n = 100$ observations sampled from the same distribution and compute the SFIMP values according to Eq. \eqref{eq:shap2}.
Panel (a) of Fig. \ref{fig:shapley} displays how the SFIMP measure distributes the total explainable performance among all features and shows the proportion of explained importance for each feature.
We repeat the experiment $500$ times on different test sets of equal size and additionally compute the difference-based and ratio-based PFI.
The results are shown in panel (b) of Fig. \ref{fig:shapley}.
For the linear model without interaction effects, the calculated importance of all three features is equal (median ratio of 1).
For all other models, we obtained a higher importance for the interacting features, indicating that these models were able to grasp the interaction effect.
However, as permuting a feature destroys any interaction with that feature, the PFI values of a feature will also include the importance of any interaction with that feature.
Thus, the importance of the interaction between $X_1$ and $X_2$ is contained in the PFI value for feature $X_1$ as well as in the PFI value for feature $X_2$.
This will overestimate the importance of $X_1$ and $X_2$ with respect to $X_3$ since $X_1$ and $X_2$ share the same interaction.
In panel (b), we thus show the ratio of the importance values with respect to $X_3$.
The results suggest that the difference-based PFI considers $X_1$ and $X_2$ twice as important as $X_3$ as the median ratio is around 2.
In contrast, the median ratio of SFIMP is around 1.5 as the importance of the interaction is fairly distributed among $X_1$ and $X_2$.

\begin{knitrout}\small
\definecolor{shadecolor}{rgb}{0.969, 0.969, 0.969}\color{fgcolor}\begin{figure}[ht]
\includegraphics[width=\maxwidth]{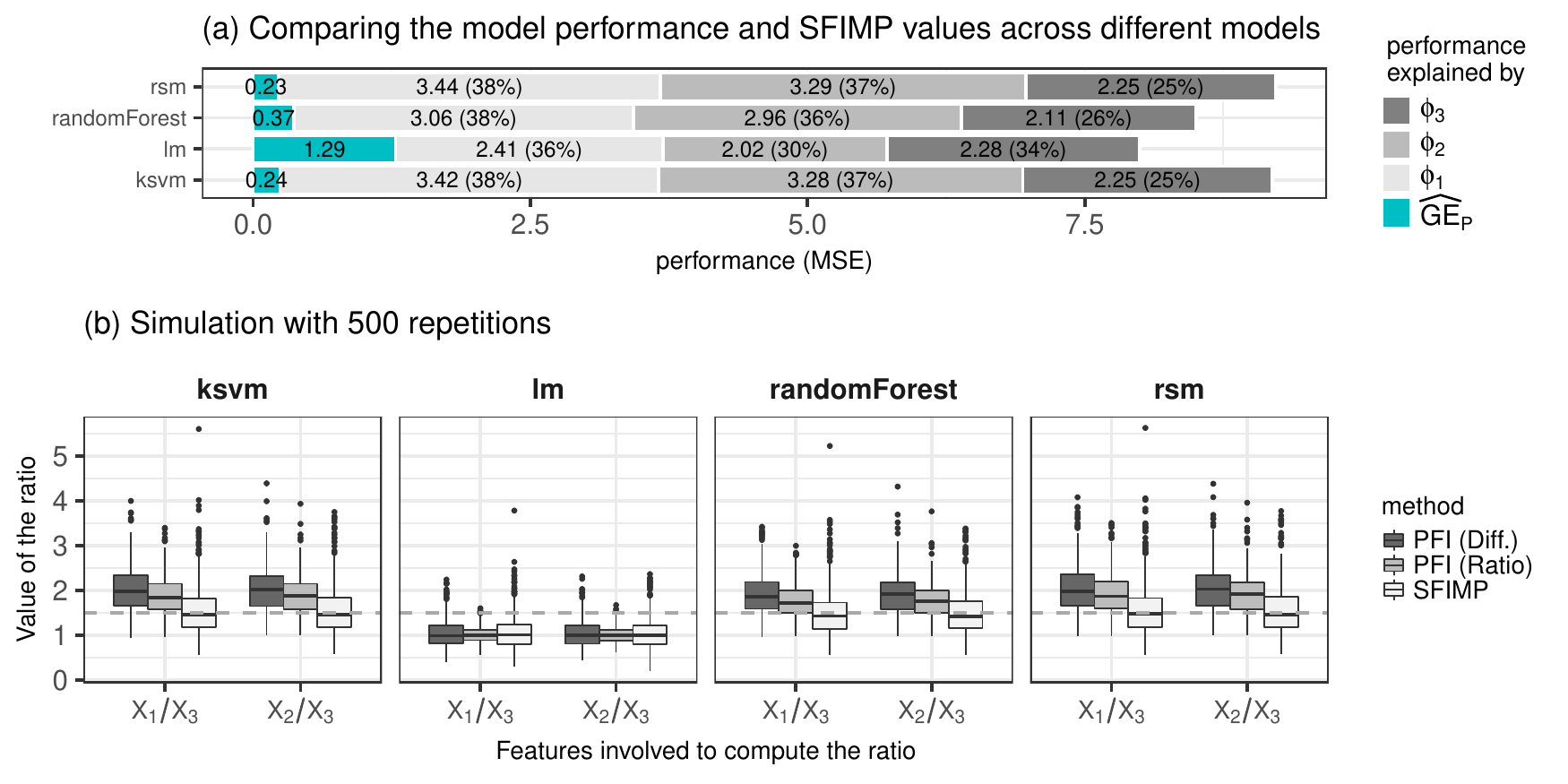} \caption{Panel (a) shows the results of a single run, consisting of sampling test data and computing the importance on the previously fitted models. The first numbers on the left refer to the model performance (MSE) using all features. The other numbers are the SFIMP values which sum up to the total explainable performance $\w[(P)]$ from Eq. \eqref{eq:performancegain}. The percentages refer to the proportion of explained importance. Panel (b) shows the results of 500 repetitions of the experiment. The plots display the distribution of ratios of the importance values for $X_1$ and $X_2$ with respect to $X_3$ computed by SFIMP, by the difference-based PFI, and by the ratio-based PFI.}\label{fig:shapley}
\end{figure}

\end{knitrout}

\subsection{Application on Real Data}

We demonstrate our graphical tools on the Boston housing data, which is publicly available on OpenML \citet{vanschoren2014openml} with data set ID 531.
The data set contains 13 features that may affect the median home price of 506 metropolitan areas of Boston.
We used the \texttt{OpenML} R package \citet{Casalicchio2017} and created the OpenML task with ID 167147 containing a holdout split ($\frac 2 3$ vs. $\frac 1 3$) for training a random forest and producing the PI and ICI plots on the test set.

Row (1) of Table \ref{tab:imp} shows the global PFI values of all features.
They are estimated using Eq. \eqref{eq:pfiest} by taking into account all $166 \cdot 166$ points of the test data.
Fig. \ref{fig:piplot} shows the corresponding PI and ICI curves for the two most important features (LSTAT and RM).
They visualize which regions of each feature and which observations have a high impact on the computed PFI values on a global and local level, which follows from the relation in Eq. \eqref{ref:relationpfi}.

PI plots visualize the expected change in performance at each position of the abscissa.
An expected change close to zero across the whole range of the feature values suggests an unimportant feature.
The PI plot of LSTAT in Fig. \ref{fig:piplot} suggests that the feature is more important if $\text{LSTAT} < 10$.
For illustration purposes, we omit all observations for which $\text{LSTAT} \geq 10$ and recompute the conditional PFI values, which are displayed in Row (2) of Table \ref{tab:imp}.
The resulting conditional PFI values are smaller, i.e., excluding observations for which $\text{LSTAT} \geq 10$ makes the LSTAT feature less important.
Note that omitting observations change the empirical distribution of the features and thus also affects the importance of other features when the PFI values are recomputed.

ICI curves additionally reveal the most (and the least) influential observations for the feature importance by considering their integral (see highlighted lines in Fig. \ref{fig:piplot}).
We can, for example, omit observations with a negative ICI curve integral.
In our test set, we observe $18$ of $166$ ICI curves with a negative integral for the LSTAT feature.
These observations have a negative impact on the global PFI according to the relation in Eq. \eqref{ref:relationpfi}.
We omit them and recompute the PFI values.
The results are listed in row (3) of Table \ref{tab:imp} and show an increased PFI value for LSTAT.

\begin{table}[ht]
\centering
\caption{PFI values calculated for a random forest trained on the Boston housing training set and using the MSE on the test data. The PFI values in row (1) are based on all observations from the test set, in row (2) on a subset where $\text{LSTAT} < 10$ and in row (3) after removing observations having a negative ICI integral.}
\label{tab:imp}
\begin{tabular}{c|c|c|c|c|c|c|c|c|c|c|c|c|c}
  \hline
 & LSTAT & RM & NOX & DIS & CRIM & PTRATIO & AGE & INDUS & TAX & RAD & B & ZN & CHAS \\
  \hline
(1) & 32.0 & 15.6 & 3.9 & 2.7 & 2.6 & 2.2 & 1.2 & 1.0 & 1.0 & 0.8 & 0.8 & 0.1 & 0.1 \\
  (2) & 10.4 & 29.6 & 1.5 & 3.3 & 0.8 & 2.3 & 0.8 & 0.5 & 1.2 & 1.1 & 0.6 & 0.2 & 0.2 \\
  (3) & 35.3 & 17.0 & 4.3 & 2.4 & 2.5 & 2.5 & 1.1 & 1.2 & 0.8 & 0.9 & 0.8 & 0.1 & 0.1 \\
   \hline
\end{tabular}
\end{table}

\begin{knitrout}\small
\definecolor{shadecolor}{rgb}{0.969, 0.969, 0.969}\color{fgcolor}\begin{figure}[ht]
\includegraphics[width=\maxwidth]{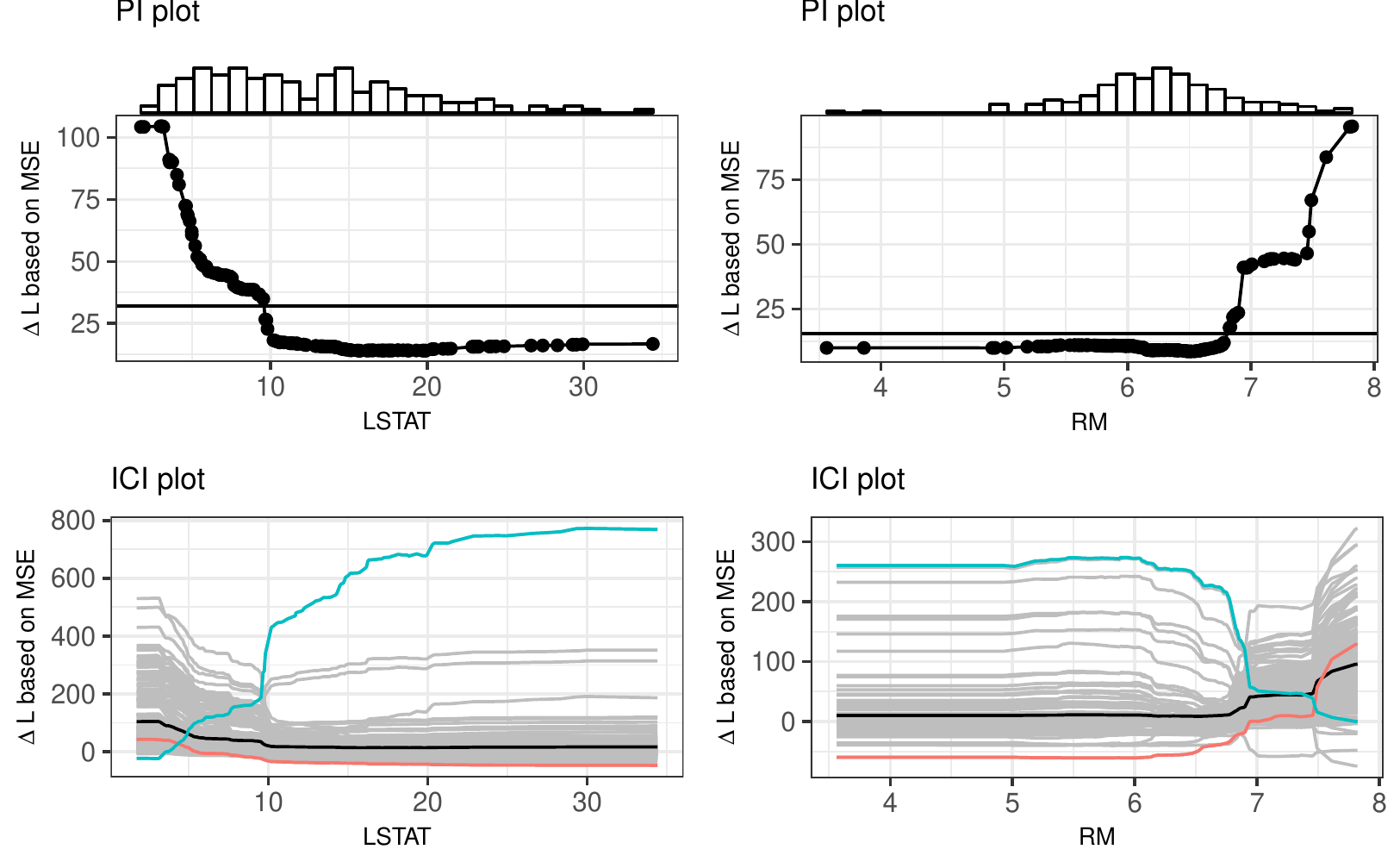} \caption[PI and ICI plots for a random forest and the two most important features of the Boston housing data (LSTAT and RM)]{PI and ICI plots for a random forest and the two most important features of the Boston housing data (LSTAT and RM). The horizontal lines in the PI plots represent the value of the global PFI (i.e., the integral of the PI curve). Marginal distribution histograms for features are added to the PI margins. The ICI curve with the largest integral is highlighted in green and the curve with the smallest integral in red.}\label{fig:piplot}
\end{figure}

\end{knitrout}

\section{Conclusion and Future Work}

It is essential for practitioners to peek inside black box models to get a better understanding of how features contribute to model predictions or how they affect the model performance.
Model-agnostic visualization methods can simplify this task tremendously.
Regarding the feature importance, the PI and ICI curves are a convenient choice for visualizing how features affect model performance.
We demonstrated how to disaggregate the global PFI into its individual local PFI components, which enabled us to visualize the feature importance on a local and global level.
It also allows practitioners to analyze and compare the feature importance across different groups of observations in the data, e.g., by subsetting the data according to other feature values and computing a conditional feature importance similar to \citet{Strobl2008} on the subsetted data which may reveal interactions.
Another interesting aspect, which we leave for future work, is aggregating the local feature importances of individual observations (i.e., the integral of ICI curves) across different features to obtain a measure for the importance of individual observations.
This could be used to find clusters of observations in the data that were important for the model performance similar to \citet{lundberg2018consistent}, but based on feature importance rather than feature effects.
Furthermore, it is also possible to disaggregate the Shapley feature importance introduced in Section \ref{sec:shapley-importance} and produce plots similar to Shapley dependency plots that were recently introduced in \cite{lundberg2018consistent}, but we leave this for future work.
Our proposed methods serve as an evaluation tool that is applied to a data set \textit{after} a model has been fitted.
As a consequence, our methods can be used to either assess the feature importance based on the ``in-sample performance'' or based on the ``out-of-sample performance'' of a fitted model.
In the former case, the same data could be used to fit the model and to calculate the quantities involved in the definition of our methods.
We focused on the latter case with independent test data.
However, we could also investigate the variability introduced by the estimation of the model itself via resampling and plot or aggregate the resulting set of quantities.

\section*{Acknowledgments}
This work is funded by the Bavarian State Ministry of Education, Science and the Arts in the framework of the Centre Digitisation.Bavaria (ZD.B).

\bibliographystyle{splncs04}
 \newcommand{\noop}[1]{}

\end{document}